\documentclass[]{acmart}
\AtBeginDocument{%
  }

\setcopyright{cc}
\copyrightyear{2018}
\acmYear{2018}
\acmDOI{XXXXXXX.XXXXXXX}
\acmConference[Conference acronym 'XX]{Make sure to enter the correct
  conference title from your rights confirmation email}{June 03--05,
  2018}{Woodstock, NY}
\acmISBN{978-1-4503-XXXX-X/2018/06}

\begin{document}

\title{The Internet of Large Language Models}
\subtitle{An Orchestration Framework for LLM Training and Knowledge Exchange Toward Artificial General Intelligence}
\author{Wilson Wei}
\email{w@eurexa.ai}
\affiliation{%
  \institution{EureXa Labs}
  \country{Singapore}
}
\author{Nicholas Chen}
\email{nok_chan@pku.edu.cn}
\affiliation{%
  \institution{Peking University}
  \country{China}
}
\author{Yuxuan Li}
\email{yuxuan.li@uwaterloo.ca}
\affiliation{%
  \institution{University of Waterloo}
  \country{Canada}
}

\renewcommand{\shortauthors}{EureXa Labs}

\begin{abstract}
This paper explores the multi-dimensional challenges faced during the development of Large Language Models (LLMs), including the massive scale of model parameters and file sizes, the complexity of development environment configuration, the singularity of model functionality, and the high costs of computational resources. To address these challenges, this paper proposes three core technical solutions: \textbf{LLM sharing protocol},  \textbf{LLM universal environment framework}, and  \textbf{Agent optimal path module}. To solve the computational resource constraints in the early stages of research, we further innovatively propose a \textbf{\textit{joint mining}} mechanism, achieving bilateral value sharing between computing power providers and model designers, including breakthrough rewards for optimal model paths and long-term profit distribution, thereby providing researchers with cost- optimized computational resource support and promoting the continuous development of LLM research and applications.
\end{abstract}



\begin{teaserfigure}
\includegraphics[width=\textwidth]{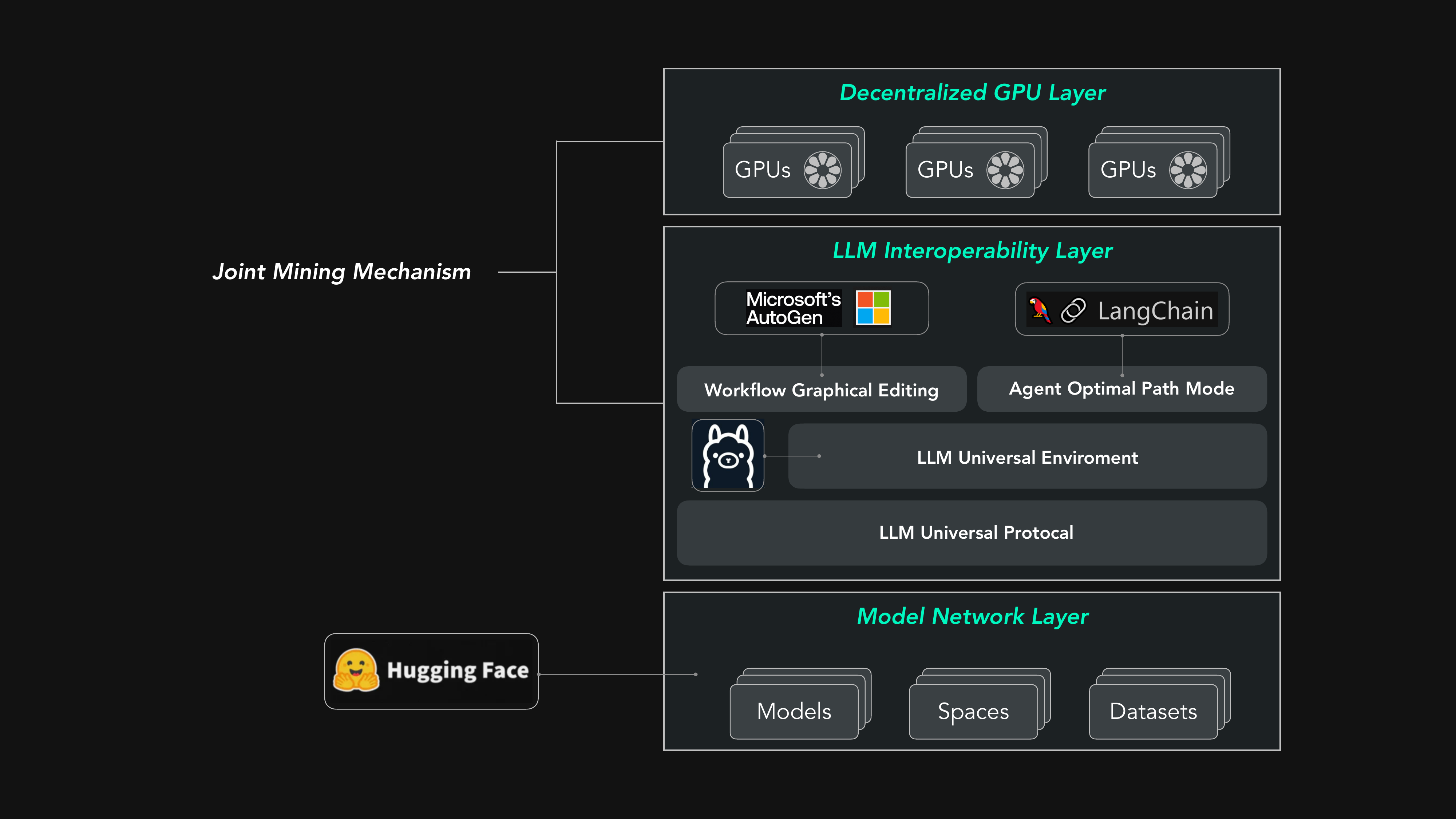}
\caption{The overview of our proposed framework.}
\Description{figure description}
\label{teaser}
\end{teaserfigure}

\maketitle

\section{Introduction}

Recently, Large Language Models (LLMs) have demonstrated their powerful semantic understanding capabilities for text and multimodal information~\citep{zheng2024video3dllm, zhu2024llava3d,brown2020llm, touvron2023llama, sun2021ernie3,chowdhery2022,gao2024cat3d,bommasani2022}. LLMs have demonstrated strong problem-solving abilities across various domains, and are becoming the foundational building blocks for the development of general-purpose AI agents or Artificial General Intelligence (AGI)~\citep{dollinger2024,lin-etal-2022-shot,workshop2023,yang2023zhongjing,wu2023bloomberggpt}. 

However, we have noticed that there are still some important issues that need to be addressed by the AI community.

\begin{itemize}
    \item \textbf{Issue 1}. Most LLMs tend to focus on specific domains, and there is no single model that consistently performs better than all others across various tasks~\citep{jiang2023llm}. Although many studies have explored the cooperation among LLMs~\citep{rame2023,wan2024,shnitzer2023large, wang2023learning,ouyang2022traininglanguagemodelsfollow,daheim2024model}, these frameworks can only accommodate a limited number of LLMs, similar to the Local Area Network (LAN) in computer networking. Can we create an  \textbf{\textit{Internet} of Large Language Models} that allows for the free transfer of knowledge among any LLMs?
    \item \textbf{Issue 2}. Given the large number of parameters and the substantial file sizes of LLMs~\citep{brown2020llm}, and the complexity on configuring development environments for LLMs, could we provide developers with a convenient model sharing and rapid environment configuration solution?
\end{itemize}
 
In the AI community, several initiatives such as Hugging Face~\citep{huggingface}, Ollama~\citep{ollama}, AutoGen\citep{wu2023autogen}, and Langchain~\citep{langchain} have been undertaken to address the aforementioned challenges. Hugging Face serves as a significant model-sharing platform, hosting a vast array of open-source models and datasets for machine learning. Ollama facilitates the local configuration and running of large models through its excellent environment design. AutoGen, an open-source framework by Microsoft, assists developers in building, managing, and optimizing multi-agent systems. Langchain is a framework for developing applications driven by language models, providing capabilities for building workflows as well as supporting the combination of agents. 

Despite the powerful capabilities of these tools, mastering and configuring the entire toolkit requires not only extensive knowledge and a strong hardware setup, but also significant patience. The high requirements for expertise and the complexity of the operations hinder the further adoption of the aforementioned tools among the general public. According to Cognitive Load Theory~\citep{sweller1988cognitive,sweller1998}, users can only process a limited amount of information at one time. Therefore, when designing tools, it is essential to consider elements such as a simple user interface, automation of repetitive tasks, and clear, visual feedback. Based on this, two new issues are arised:
\begin{itemize}
    \item \textbf{Issue 3}. Can we develop a tool that can achieve one-click operations—including environment deployment, model downloads, development ebugging, and publishing sharing?
    \item \textbf{Issue 4}. After optimizing based on the Baseline, could this tool automatically explore and combine LLMs to form the optimal Agent path?
\end{itemize}

With these issues in mind, to understand the real needs from the perspective of front-line developers and researchers in identifying workable solutions, we engaged in comprehensive discussions with 63 experts working in fields such as Large Language Models, Reinforcement Learning, Robotics, and Computer Graphics, including employees of leading companies in the industry and graduate students. Through the interactions, we identified another pressing issue:  
\begin{itemize}
    \item\textbf{Issue 5}. The computational costs for training models are excessively high~\citep{cottier2024, huang2024, Shi2024}.  Can we significantly reduce the cost of training by designing a new distributed training framework?
\end{itemize}

These issues led us to realize that the AI-driven industrial revolution of this century still has a considerable distance to travel. From the development of tools and replication of baselines to the implementation of upper-layer applications, each step encounters significant obstacles, whether they be in technical requirements, time investment, or financial costs. These are challenges that the AI community must address and overcome collectively.

Considering the above issues, in this paper, we propose a new framework for LLM training and knowledge exchange, namely \textbf{\textit{The Internet of LLM}}. Figure~\ref{teaser} shows the overview of the framework. We implemented four innovations in the Internet of LLM, with some based on secondary development of Langchain and Ollama:
\begin{enumerate}
    \item  \textbf{LLM Sharing Protocol}: Given the substantial need for LLMs in using, constructing Workflows, and developing Agents, the rapid transfer of LLMs across different regions presents a significant challenge. To address this, we devised a universal LLM model protocol to facilitate one-click integration and swift sharing.
    \item  \textbf{LLM Universal Environment}: Diverse environments pose numerous adaptation challenges for developers and users. To reduce entry barriers, we established a unified platform for development, training, and execution, thereby minimizing the time developers spend on environment setup, version control, and troubleshooting.
    \item  \textbf{Agent Optimal Path}: When handling complex tasks, the system continually selects and combines models, conducts joint training of multiple models, and evaluates and provides feedback on the results. Due to the time-intensive nature of this process, we employed parallel computing techniques to expedite the search for the optimal pathway that meets the current requirements.
    \item  \textbf{Joint Mining}: To alleviate the initial computational costs for researchers, computing power providers can contribute by offering computational resources. In return, they share two benefits with model designers: breakthrough rewards for optimal model pathways and long-term returns from the models. This arrangement enables researchers to access discounted or complimentary computational support.
\end{enumerate}

Our project will contribute to the AI community and the broader human society in the following aspects:
\begin{itemize}
    \item \textbf{Facilitating Knowledge Sharing and Collaboration:} The establishment of this framework provides an open and shared platform for researchers and developers worldwide, fostering knowledge exchange and collaboration across different regions and institutions. Such a collaborative model contributes to the aggregation of global intelligence and resources, jointly advancing the development of general artificial intelligence, and accelerating its application and popularization across various societal domains.
    \item \textbf{Lowering Technical Barriers and Promoting Widespread Adoption:} By simplifying operational procedures and reducing computational costs, the framework enables more non-specialists to access and utilize general artificial intelligence technologies, thereby expanding its audience. This facilitates an increase in societal awareness and acceptance of general artificial intelligence, creating favorable conditions for its broader dissemination and application at various social levels.
    \item \textbf{Promoting Industrial Ecosystem Development:} The establishment and enhancement of this framework attract more enterprises, research institutions, and developers to participate in the general artificial intelligence industry chain, fostering a virtuous cycle within the ecosystem. This, in turn, stimulates the development of related hardware, software, and data services industries, propelling the prosperity and economic growth of the entire general artificial intelligence industry.
    \item \textbf{Promoting the Development of Greener AI.} This framework enhances the efficiency of computational resource utilization through the optimization of training processes and resource-sharing mechanisms. By reducing the training costs of LLMs, it lowers energy consumption, thereby decreasing carbon emissions. This promotes a shift in the demand for computational resources towards more environmentally friendly and sustainable directions.
\end{itemize}

\section{Case Study of Related Work}
\subsection{Ollama}
Ollama~\citep{ollama} is a platform specifically designed for local deployment, running, and managing large language models (like LLaMA). It adopts Docker-like operations, allowing non-professional users to easily manage and use these complex models without relying on cloud services and complex infrastructure configurations.

Features of Ollama: 
\begin{enumerate}
    \item  \textbf{Independent Environment}: Ollama provides a simple and convenient deployment method for large language models through Docker containers, effectively lowering technical barriers and saving users significant configuration time and effort.
    \item   \textbf{Lightweight and Scalability}: The framework has low resource consumption and supports flexible configuration adjustments on demand, adaptable to projects and hardware environments of different scales.
    \item   \textbf{Pre-built Model Library}: Includes a series of pre-trained models that users can use directly without training themselves
    \item  \textbf{Multi-platform Support}: Full support for macOS, Linux, and Windows systems, allowing users to use it seamlessly on any mainstream operating system
    \item  \textbf{Command Line Tools}: Provides a streamlined command-line interface to start services and supports custom environment variables to meet personalized needs
\end{enumerate}

\subsection{Hugging Face}
Hugging Face~\citep{huggingface} is a popular open-source community and platform dedicated to advancing open-source natural language processing and machine learning. As the GitHub of machine learning, the platform offers numerous open-source models, datasets, and applications, equipped with comprehensive documentation and active community support, making it convenient for developers to learn and use.

Features of Hugging Face:
\begin{enumerate}
\item  \textbf{Transformers Library}: This library provides thousands of pre-trained models, supporting various natural language processing tasks such as text generation, sentiment analysis, and named entity recognition. It is fully compatible with mainstream deep learning frameworks like PyTorch and TensorFlow, allowing users to flexibly choose their development environment.
\item   \textbf{Model Sharing Platform}: Hugging Face provides functionality for users to upload, download, and share machine learning models and datasets, building a vibrant community. Users can test models directly on the platform without local deployment, greatly improving development efficiency.
\item \textbf{ Multimodal Support}: Hugging Face supports not only text processing but also image, audio, and other multimodal tasks, significantly expanding its application scenarios.
\item   \textbf{Datasets Library}: Integrates comprehensive dataset management functionality, enabling users to conveniently load, process, and share datasets, effectively supporting model training and evaluation work.
\item  \textbf{Community and Documentation}: Hugging Face has an active developer community where members actively share experiences, tutorials, and models, promoting technical exchange and collaboration.
\end{enumerate}

\subsection{AutoGen}
AutoGen~\citep{wu2023autogen} is an open-source framework developed by Microsoft Research team, focusing on simplifying the creation and management of multi-agent systems, particularly for Large Language Model (LLM) applications. It provides a unified multi-agent dialogue framework that enables multiple agents to collaborate through conversation to solve complex tasks.

Features of AutoGen:
\begin{enumerate}
\item  \textbf{Multi-agent Collaboration}: AutoGen supports dialogue interaction between multiple agents, allowing them to collaboratively handle complex tasks. These agents can be customized entities integrating LLMs, tools, or human input, capable of flexibly addressing various needs.
\item  \textbf{Workflow Customization}: Developers can customize agents according to specific requirements, creating intelligent entities with specific functionalities. This modular design enables developers to build diverse LLM applications suitable for different domains.
\item  \textbf{Dynamic Dialogue Mode}: AutoGen supports dynamic dialogue mechanisms, allowing agents to flexibly adjust based on real-time conversation content. This feature is particularly suitable for handling interaction scenarios in complex applications that cannot be preset.
\item  \textbf{Human-Machine Collaboration}: AutoGen supports human participation mechanisms, incorporating human feedback at crucial points to optimize decision-making processes through configurable human input modes.
\item  \textbf{Integration and Extensibility}: AutoGen is equipped with modules such as model, skill, and agent, enabling seamless integration with various tools and APIs, allowing developers to easily access external resources. Additionally, users can flexibly extend and combine different agents to meet specific needs.
\end{enumerate}

\subsection{LangChain}

LangChain~\citep{langchain} represents an open-source framework specifically engineered for the development of sophisticated applications built upon Large Language Models (LLMs). This framework provides comprehensive tools and modules that enable developers to seamlessly integrate language models with external data sources, APIs, and user interfaces, facilitating the creation of robust applications.
The ecosystem primarily comprises two essential tools: LangGraph and LangSmith, which collectively facilitate the construction and optimization of LLM applications. The key functionalities are delineated as follows:
\begin{itemize}
    \item \textbf{LangChain}~\citep{langchain}: Serving as the foundational framework, it implements a structured methodology for LLM operations. Through its chain-based workflow mechanism, developers can systematically orchestrate multiple processing steps to efficiently execute complex tasks.
    \item \textbf{LangGraph}~\citep{langgraph2024}: This LangChain extension introduces graphical architecture, enabling developers to construct sophisticated multi-agent systems. Through state management and cyclic workflows, it ensures coordinated agent operations while maintaining contextual coherence.
    \item \textbf{LangSmith}~\citep{langsmith2024}: This specialized monitoring and debugging utility facilitates the tracking of model inputs and outputs, enabling prompt issue identification and resolution. It delivers comprehensive testing and evaluation capabilities, effectively supporting application optimization throughout the development lifecycle.
\end{itemize}

\subsubsection{LangChain Features}
\begin{enumerate}
    \item \textbf{Contextual Awareness}: Connects language models to contextual sources (such as prompts and examples), enabling applications to comprehend and respond to user inputs effectively.
    \item  \textbf{Reasoning Capabilities}: Leverages language models for inference, generating responses and actions based on contextual understanding.
    \item  \textbf{Modular Architecture}: Offers composable tools and integrations, facilitating the development of sophisticated applications.
    \item  \textbf{Templates and Pre-built Chains}: Provides deployable reference architectures and ready-to-use chains, expediting development initiation.
\end{enumerate}

\subsubsection{LangGraph Features}
\begin{enumerate}
    \item  \textbf{Cyclic Flow Support}: Enables the definition of processes incorporating loops, suitable for applications requiring memory and contextual reasoning.
    \item  \textbf{State Management}: Facilitates information storage and retrieval across multiple steps, ideal for tracking conversation or game states.
    \item  \textbf{Multi-agent Support}: Enables interaction between multiple agents, applicable to collaborative or competitive scenarios.
    \item  \textbf{Usability and Flexibility}: Provides intuitive APIs, ensuring accessibility for newcomers.
\end{enumerate}

\subsubsection{LangSmith Features}
\begin{enumerate}
 \item  \textbf{Comprehensive Pipeline Support}: Delivers end-to-end support from prototyping through production stages.
 \item \textbf{Debugging and Monitoring Capabilities}: Assists developers in swift problem identification and resolution, enhancing application quality.
 \item \textbf{LangChain Integration}: Seamlessly integrates with the LangChain framework, enabling efficient application debugging and optimization.
\end{enumerate}

\section{System Architecture}
The entire system is divided into three layers. Model Network Layer, LLM Interoperability Layer, and Decentralized GPU Layer.
\begin{itemize}
    \item \textbf{Model Network Layer}. This layer supports the integration of various models (Models), applications (Spaces), and datasets (Datasets), with LLMs being a crucial model category. We have established mirror sites at global nodes to ensure fast data transmission for designers and users. Additionally, this layer is compatible with Hugging Face interfaces, thus enriching the variety and quantity of available models.
    \item \textbf{LLM Interoperability Layer}. This layer contains four core components: LLM universal protocol, LLM universal environment, Workflow graphical editor, and Agent optimal path module. They respectively provide LLM sharing transmission protocol, training and testing environment, graphical interface for LLM Workflow construction, and a functional module for autonomous exploration of optimal Agent paths.
    \item \textbf{Decentralized GPU Layer}. We will connect to existing GPU computing platforms and record the benefits generated from training models under the computing power provider's ID. Through a ``joint mining'' mechanism, computing power providers can negotiate benefit distribution ratios with model trainers, thereby achieving computing power investment.
\end{itemize}

\subsection{Model Network Layer}
\subsubsection{Multi-model Compatibility}
The Model Network Layer supports various types including deep learning models, traditional machine learning models, and pre-trained models, with a primary focus on Large Language Models (LLM). While handling large-scale model parameters and data volumes, this layer can also seamlessly integrate models from multiple fields such as natural language processing and computer vision.

\subsubsection{Modular Design}
The system adopts a modular architecture, encapsulating each model as an independent module, similar to GitHub's repository management approach. This design facilitates individual updates and maintenance of models, while tracking change history through version control systems, ensuring project traceability and stability.

\subsubsection{Unified Interface}
The system provides a unified API interface to simplify model calling processes. Developers can complete model loading, inference, and evaluation through this standardized interface without needing to understand underlying implementation details. This not only improves development efficiency but also makes the integration of various model platforms more convenient.

\subsubsection{Integration and Deployment}
The system can seamlessly integrate with existing development toolchains and deployment platforms. Following Hugging Face's model library design, the Model Network Layer provides convenient model import and export functionality, supporting various deployment environments including local servers, cloud platforms, and edge devices.

\subsection{LLM Interoperability Layer}
\begin{figure*}[htbp]
\includegraphics[width=.75\linewidth]{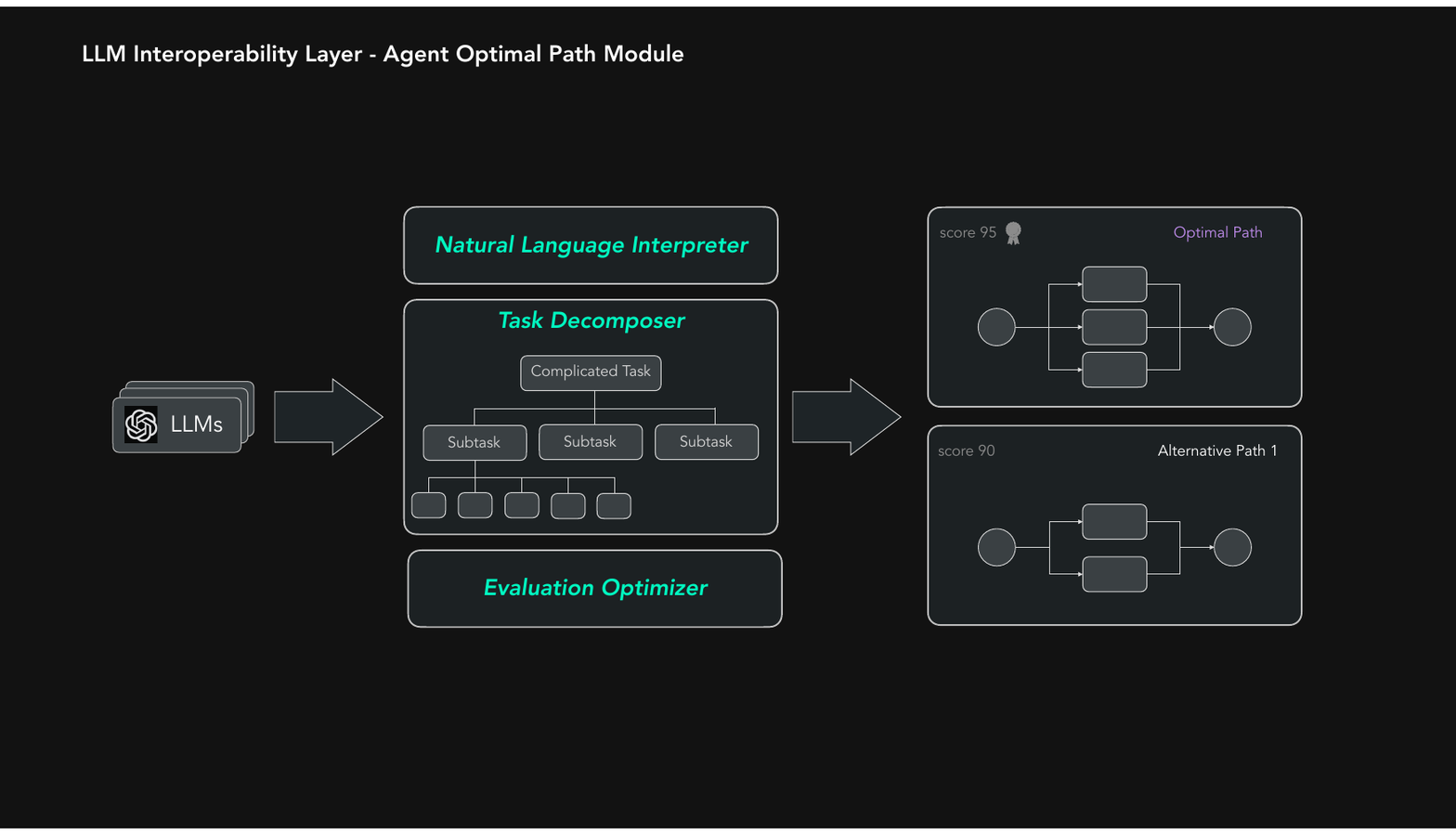}
  \caption{The illustration of LLM Interoperability Layer.}
  \Description{}
  \label{fig:dimensions}
\end{figure*}
\subsubsection{LLM Sharing Protocol}
The LLM sharing protocol includes several key components, collectively building a robust and flexible communication framework.
\begin{itemize}
    \item \textbf{Data Format Standardization}. Adopting unified data structures (JSON and Protocol Buffers) establishes a standard foundation for data exchange between models. The system uses a unified JSON request format, including clearly defined request types, parameter configurations, and contextual information, ensuring consistency in data interactions.
    \item \textbf{Communication Interface Specification}. Establishes standardized API structures, integrating RESTful API and gRPC request methods, equipped with complete endpoint definitions and security authentication mechanisms, implementing standardized model calling processes.
    \item \textbf{Version Control Mechanism}. Adopts systematic protocol version management, ensuring seamless collaboration between different protocol versions through automated version incrementation and backward compatibility guarantees. This multi-layered architectural design significantly enhances system communication efficiency, scalability, and maintainability. 
\end{itemize}

\subsubsection{LLM Universal Environment}
The LLM universal environment provides a unified and efficient runtime and development environment, specifically designed for Large Language Models (LLMs) with large parameter counts and model sizes. This environment integrates Ollama technology, not only simplifying LLM deployment and management processes but also optimizing resource utilization to ensure high performance and scalability.

\begin{itemize}
    \item \textbf{Containerization Support}. The system employs container technologies like Docker and Kubernetes, encapsulating various LLMs in independent containers to ensure environment consistency and portability. Ollama provides optimized large model container images with built-in necessary dependencies and configurations, supporting rapid deployment and elastic scaling. This containerization approach makes LLM deployment more modular and facilitates cross-environment migration and management.
    \item \textbf{GPU Optimization Scheduling}. Ollama integrates intelligent scheduling algorithms that can dynamically allocate GPU resources based on model demands and resource conditions, maximizing computational efficiency. For example, the system automatically allocates more GPU instances during peak periods to meet concurrent request demands.
    \item \textbf{Memory and Storage Optimization}. 
    The system employs distributed storage and memory management technologies to ensure efficient loading and access of large model data. Through compression techniques and memory paging mechanisms, it effectively reduces memory usage and improves data transfer speeds.
    \item \textbf{Automatic Scaling}. Based on Kubernetes' auto-scaling functionality, the universal environment can automatically adjust the number of model instances according to real-time load, maximizing resource utilization while effectively controlling operational costs.
\end{itemize}

\subsubsection{Workflow Graphical Editor}
The workflow graphical editor is an integrated visual tool specifically designed to simplify the design, configuration, and management of complex Large Language Model (LLM) workflows. By integrating advanced tools like AutoGen and LangGraph, this editor not only optimizes user experience but also significantly enhances workflow flexibility and scalability. Below we will detail the editor's core components and functions.
Users only need to drag and drop predefined nodes (including input, processing, output nodes, etc.) onto the canvas to intuitively build workflows. Each node represents specific operations or steps and supports various LLM tasks such as text generation, translation, and sentiment analysis.
\subsubsection{Agent Optimal Path Module}
The Agent optimal path module builds an intelligent and efficient task optimization system by integrating key components such as natural language interpreters, LLM planners, reflection and improvement, memory-enhanced planning, collaborative training, and evaluation. These components work together to ensure the system can accurately understand user requirements, formulate optimal execution plans, and continuously improve through feedback and optimization.
This content will be detailed and analyzed in Section ~\ref{agent_optimal_path}.

\subsection{Decentralized GPU Layer}
The decentralized GPU layer adopts a distributed architecture, composed of GPU nodes distributed across different data centers and geographical locations. Although nodes operate independently, they achieve collaborative work through efficient network connections.
Each node cluster is equipped with multiple GPUs, possessing independent computing and storage capabilities to handle large-scale parallel computing tasks. We adopt a heterogeneous computing architecture, combining general-purpose CPUs with specialized processors. Compared to traditional homogeneous architectures, this design can more effectively handle diverse tasks. Different types of processors have their respective strengths, and through reasonable task allocation, system overall performance can be significantly improved. Specialized processors excel in specific tasks, not only delivering superior performance but also effectively reducing system energy consumption.

The decentralized GPU layer adopts a distributed architecture, composed of GPU nodes distributed across different data centers and geographical locations. Although nodes operate independently, they achieve collaborative work through efficient network connections.
Each node cluster is equipped with multiple GPUs, possessing independent computing and storage capabilities to handle large-scale parallel computing tasks. We adopt a heterogeneous computing architecture, combining general-purpose CPUs with specialized processors. Compared to traditional homogeneous architectures, this design can more effectively handle diverse tasks. Different types of processors have their respective strengths, and through reasonable task allocation, system overall performance can be significantly improved. Specialized processors excel in specific tasks, not only delivering superior performance but also effectively reducing system energy consumption.

\section{LLM Sharing Protocol}
\subsection{Standardized Model Integration Protocol}
The Standardized Model Integration Protocol (SMIP) establishes an efficient, reliable, and flexible model integration framework. This framework preserves complete datasets, parameters, and models, provides one-click upload and download functionality, supports multi-platform compatibility, implements data format standardization, standardizes communication interfaces, and includes version control mechanisms. The implementation of SMIP not only significantly enhances the convenience of model management and integration but also promotes the healthy development of the AI ecosystem and multi-party collaboration, providing strong support for continuous innovation in research and applications.

\subsubsection{Complete Preservation of Datasets, Parameters, and Models}
SMIP ensures complete preservation and lossless transmission of datasets, model parameters, and model structures during the model integration process. Regardless of data format or model complexity, the protocol ensures that all critical information remains intact and unaltered during migration, thus maintaining the original performance and accuracy of models. This mechanism provides researchers and developers with a reliable foundation, allowing their work to flow freely between different platforms without concerns about data or model integrity.

\subsubsection{One-Click Upload and Download}
SMIP provides streamlined and efficient one-click upload and download functionality, greatly simplifying the model integration process. Users only need to click the upload button through a unified interface to transfer local models, datasets, and parameters to the target platform in one go, without manual configuration or step-by-step operations. Similarly, the one-click download feature allows users to quickly download models and related resources from the target platform to their local environment for subsequent use and deployment. This convenient operation not only enhances user experience but also lowers technical barriers, making model sharing and application more efficient.

\subsubsection{Multi-Platform Compatibility}
SMIP is designed to be compatible with multiple mainstream platforms, ensuring interoperability between different ecosystems. Specifically, the protocol supports seamless integration with platforms like Ollama and LangGraph, allowing users to easily import and export models in these environments. Furthermore, through modular design, SMIP can extend support to more third-party platforms, meeting users' diverse integration needs. This broad compatibility allows users to focus on model development and application without worrying about platform differences.

\subsubsection{Data Format Standardization}
To ensure smooth data exchange between different platforms, SMIP defines unified data format standards. The protocol adopts common data formats like JSON and YAML to ensure consistency of model descriptions and parameter configurations across different systems. The standardized format supports extensions, allowing custom fields to be added based on specific needs to meet complex model description requirements. Through data format validation mechanisms, SMIP ensures uploaded and downloaded data complies with protocol specifications, preventing integration failures due to format errors and guaranteeing data exchange reliability and consistency.

\subsubsection{Communication Interface Specification}
SMIP establishes unified communication interface specifications to ensure efficient and reliable interaction between platforms. The protocol defines a set of RESTful API interface standards covering model upload, download, query, and update operations, greatly simplifying developers' integration work. In terms of security, the protocol employs HTTPS and OAuth security mechanisms to effectively protect data transmission security and privacy. Additionally, standardized error codes and response formats enable developers to quickly locate and resolve integration issues, thereby enhancing system stability and user trust.

\subsubsection{Version Control}
To effectively manage model iterations and updates, SMIP integrates version control mechanisms. Each model upload automatically receives a unique version identifier, facilitating tracking and management of different versions. The protocol supports version rollback functionality, allowing quick recovery to stable versions when updates encounter issues. Change logs detail the updates of each version, including feature optimizations and bug fixes, providing a clear iteration history. This mechanism not only ensures the flexibility and controllability of model development but also lays a solid foundation for team collaboration and long-term maintenance.

\subsection{LLM Output Caching}
LLM Output Caching is a mechanism for storing and managing model output results in large-scale language model applications. Through efficient output caching, the system can significantly improve response speed, reduce computational resource consumption, and optimize user experience. Below are the core aspects of LLM Output Caching.

\subsubsection{Repeated Retrieval Output Caching}
LLM Output Caching addresses resource waste from repeated queries by pre- storing responses to common queries. When users make identical or similar requests, the system retrieves results directly from cache without needing to recalculate using the model. This not only reduces response time but also achieves fast retrieval through efficient indexing mechanisms, improving overall performance. Combined with associated search engines, the system can better utilize LLM resources, ensuring excellent performance in high-concurrency and real-time application scenarios.
\subsubsection{Reducing Computational Resource Consumption}
LLM Output Caching significantly reduces computational resource requirements since identical requests don't need repeated model processing. This optimization is particularly important in cloud computing environments, effectively reducing operational costs. Through intelligent caching strategies, such as dynamic adjustments based on request frequency and response time, the system can efficiently manage resources. Storing common results not only reduces repeated calculations but also maximizes system cost-effectiveness through efficient resource management.
\subsubsection{Data Format Standardization and Compatibility}
LLM Output Caching adopts standardized data formats to ensure cross-platform compatibility. Using unified formats like JSON ensures model descriptions and parameter configurations remain consistent across different environments. Standardization not only facilitates cache management and maintenance but also supports cross-platform data exchange, enhancing system flexibility. Through indexing and search engines, cached content can be quickly accessed, improving processing efficiency while ensuring seamless integration with other data processing workflows.

\subsection{Search Engine Indexing}
Efficient LLM output caching requires intelligent management strategies, including update, invalidation, and elimination mechanisms. The system dynamically optimizes storage strategies by analyzing request patterns and cache hit rates. Using methods such as Least Recently Used (LRU) algorithms and time-based invalidation mechanisms ensures the cache maintains the most valuable content. Through preloading and warm-up mechanisms, combined with efficient indexing and search engines, the system can more intelligently predict and respond to user needs, achieving optimal utilization of cache resources.

\section{LLM Universal Environment}
\subsection{ Isolated Environment Architecture}
This section primarily draws from the design philosophies of Ollama and Anaconda. The isolated environment architecture is a core technical infrastructure designed specifically for Large Language Models (LLM) and Machine Learning (ML) applications. This architecture creates independent runtime spaces to ensure complete isolation of environments for different models and experiments, thereby enhancing system stability and reliability.

\subsubsection{Dependency Management} 
In complex machine learning ecosystems, dependency management is crucial. Our architecture provides independent runtime environments for each project, effectively resolving version conflict issues. For example, when one project requires TensorFlow 2.0 while another needs PyTorch 1.9, the system can perfectly maintain these two separate environments, ensuring smooth workflow progression. 

\subsubsection{Environment Independence}
Environment independence is the cornerstone of this architecture. Through strict environment isolation mechanisms, each project runs in its dedicated space, effectively preventing mutual interference. This not only enhances system security but also provides an ideal workspace for development and testing. Researchers can freely conduct model training and debugging in independent environments while maintaining production environment stability.

\subsubsection{Environment Migration and Collaboration}
Our architecture provides advanced environment migration and collaboration capabilities. Research teams can easily export complete environment configuration templates, enabling rapid environment replication and deployment. This greatly improves team collaboration efficiency, ensuring all members work under identical environment configurations, effectively avoiding the ``it works on my machine'' problem.

\subsubsection{Version Control System}
The version control system employs precise dependency management mechanisms and integrates mainstream package management tools. The system supports automatic dependency resolution and provides version locking functionality, ensuring reproducibility in model training processes. Through this approach, researchers can ensure the reproduction of identical experimental results at different points in time.

\subsubsection{Multi-Platform Compatibility}
Our architecture achieves exceptional cross-platform compatibility. Through unified configuration standards and environment management strategies, it ensures consistent model performance across different operating systems. Whether in a Windows development environment or on Linux production servers, models maintain the same performance levels and operational effects, greatly simplifying the deployment process.

\subsection{Docker Container Image}
This project adopts the environment management philosophy of Ollama and Anaconda in its Docker container image design. As a key technical component of Isolated Spaces, Docker container images play a central role in environment management for LLM and machine learning models. Through integrating applications and their dependencies into lightweight, portable containers, we have achieved environment standardization and consistency.
By integrating Ollama's professional environment management architecture with Anaconda's efficient package management system, we have built independent runtime environments for each LLM model, effectively avoiding dependency conflicts between models. From a technical implementation perspective, Docker images provide research teams with a comprehensive environment encapsulation solution, enabling rapid deployment, flexible scaling, precise version control, and collaborative development, significantly enhancing the practical value and applicability of Isolated Spaces.

\subsection{GPU Confidential Computing}
GPU confidential computing is a professional security technology solution designed to ensure the confidentiality and integrity of data during GPU processing. Given GPU's current core role in important fields such as artificial intelligence and machine learning, data security protection has become a key requirement. This technology provides comprehensive security protection for data processing on GPUs through integrated hardware and software solutions. 

Key features of GPU Confidential Computing:
\begin{enumerate}
    \item \textbf{Full-cycle Data Protection Mechanism}. Implements comprehensive data encryption strategies covering static storage, transmission processes, and real-time processing stages, ensuring data security throughout its lifecycle.
    \item \textbf{Professional Execution Environment}. Provides independent secure computing spaces with strict data access control, maintaining data security even in the event of system attacks.
    \item \textbf{Hardware Security Architecture}. Fully utilizes GPU hardware security features, including memory encryption, data channel protection, and physical isolation technologies, to build a multi-layered security protection system.
\end{enumerate}

\subsection{Cross-Platform Compatibility}
\subsubsection{Multi-language Development Environment}
This framework provides comprehensive multi-language support, integrating mainstream programming languages including Python, JavaScript, Java, C\#. Through standardized API interfaces and SDK toolsets, it ensures consistency and reliability in cross-language functionality implementation. Complete technical documentation and practical examples provide professional guidance support for developers, effectively promoting efficient application of framework features.

\subsubsection{Cross-platform Runtime Environment}
This framework achieves comprehensive compatibility across major operating systems including Windows, macOS, and Linux. Through deep adaptation of various system versions, including long-term support for LTS versions, it ensures a stable and reliable runtime environment. The framework fully integrates system- specific features, such as Windows system configuration management and macOS performance optimization, achieving excellent system performance and user experience.

\subsubsection{Multi-environment Deployment Solution}
This framework achieves seamless integration with mainstream cloud service platforms (AWS, Azure, Google Cloud), providing development teams with flexible deployment options. It simultaneously supports deployment requirements for local servers, proprietary data centers, and hybrid cloud architectures, fully meeting enterprise-level security and compliance standards. Built-in Docker and Kubernetes support simplifies containerized deployment processes and optimizes cross-platform migration efficiency. Integration with professional CI/CD toolchains achieves excellence in automated deployment and operations management.

\subsubsection{Modular Extension Architecture}
This framework adopts a streamlined core modular design philosophy, implementing flexible configuration and management of plugin functionality. Through standardized plugin interfaces and development standards, it ensures interoperability between modules and system stability. It supports community and third-party developer innovation contributions, building a rich plugin ecosystem. Equipped with dynamic loading and hot update mechanisms, it minimizes system maintenance downtime while continuously improving application availability and performance.

\section{Agent Optimal Path Module}
\label{agent_optimal_path}
\begin{figure*}[htbp]
\includegraphics[width=.95\linewidth]{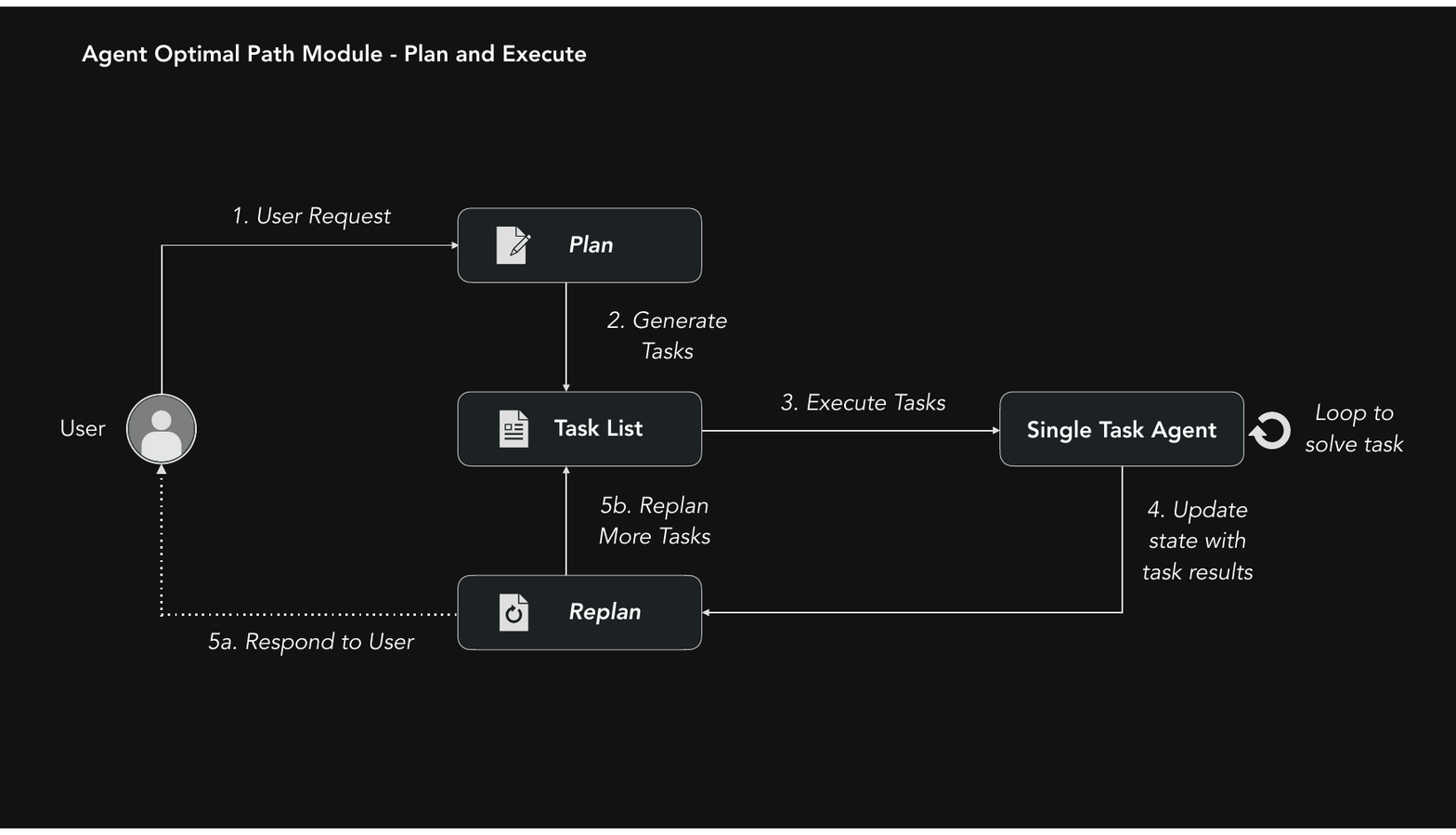}
  \caption{The illustration of how agents plan and execute tasks.}
  \Description{}
  \label{fig:dimensions}
\end{figure*}
\begin{figure*}[htbp]
\includegraphics[width=.95\linewidth]{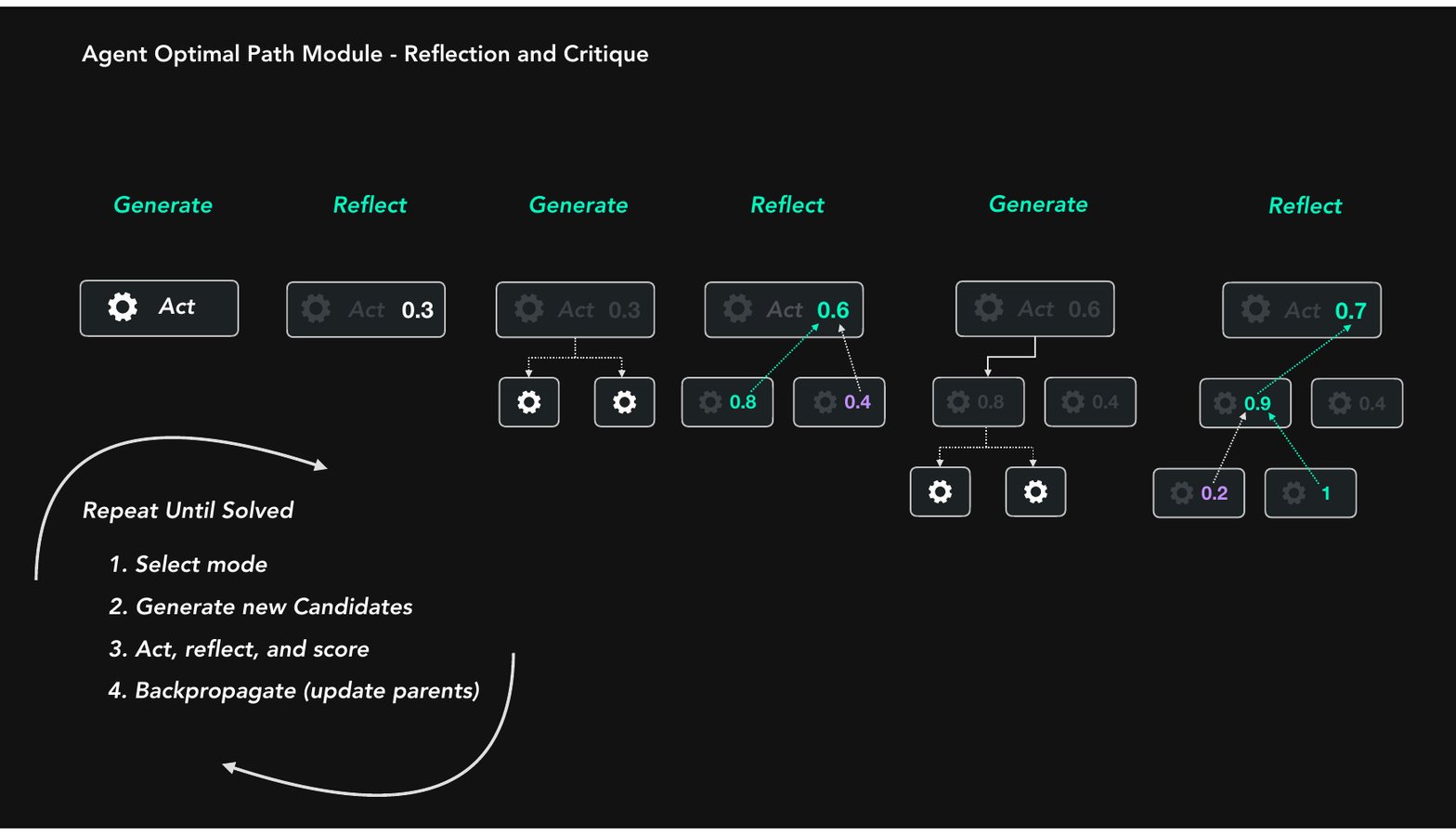}
  \caption{The illustration of how to reflect and critique in automatically finishing tasks.}
  \Description{}
  \label{fig:dimensions}
\end{figure*}
Optimal Path refers to selecting the most efficient execution path or strategy combination from multiple feasible solutions through systematic evaluation criteria and optimization algorithms when executing complex tasks or designing systems.
In building and optimizing multi-level LLM combination systems, the implementation of Optimal Path primarily relies on the collaborative operation of two core technical components: LLM Planner and Optimal Path Evaluator. LLM Planner focuses on constructing multiple LLMs into a clearly hierarchical tree structure, ensuring clear responsibilities at each level; while the Optimal Path Evaluator is responsible for real-time evaluation of generated execution paths, continuously optimizing execution efficiency at each level to enhance overall system performance. This specialized architectural design not only significantly improves the system's intelligence level but also achieves continuous optimization of optimal paths through precise parameter tuning and system training.
Based on this, Optimal Path has become a key indicator of continuous innovation and technological breakthrough in Agent systems, with importance comparable to world records in competitive fields. Against the backdrop of continuous LLM technology development, the exploration of optimal paths is not only a core element in improving system performance but also an important driving force in promoting the evolution of the overall technical architecture.

\subsection{Natural Language Interpreter}
The natural language interpreter significantly enhances system intelligence through deep integration with Large Language Models (LLM). This component specializes in precise parsing of user inputs, utilizing LLM's advanced language processing capabilities to ensure accurate understanding of user intent and context.
At the technical implementation level, the interpreter converts natural language into standardized structured data, such as JSON format, SQL queries, or API call instructions. Leveraging LLM's contextual understanding and multi-task processing capabilities, the system generates precise execution instructions and dynamically adjusts conversion strategies based on specific application scenarios.
In terms of system integration, the interpreter serves as a professional conversion interface between user requirements and system functionality. Through the integration of LLM's knowledge base and reasoning capabilities, it significantly improves the accuracy of tool invocation. For example, for a query about ``bank loan interest rate comparison,'' the system can construct professional retrieval logic, call relevant data interfaces, and present analysis results in a professional yet comprehensible manner.
In task processing, the interpreter employs advanced task decomposition strategies. Taking restaurant reservations as an example, the system breaks down complex requests into multiple execution steps, completing tasks through LLM's multi-level reasoning capabilities according to a clear execution sequence (location confirmation, seat availability check, booking confirmation).
Through systematic integration with LLM, the interpreter not only efficiently handles diverse task requirements but also continuously enhances the professionalism and efficiency of interaction experiences through ongoing optimization mechanisms (including iterative updates and validation processes).

\subsection{LLM Planner}
As the core component of task planning, the LLM Planner is specifically responsible for converting user input into structured execution plans. The system employs advanced contextual analysis technology to systematically break down complex tasks into clear execution units. Taking restaurant reservations as an example, the system follows a professional process to sequentially execute restaurant search, seat availability checks, and booking confirmation steps to ensure service quality.
During task execution, the LLM Planner demonstrates excellent dynamic adjustment capabilities. Through real-time monitoring and feedback analysis, the system can promptly optimize execution strategies and improve operational efficiency. When interfacing with external systems, LLM Planner adopts standardized interface protocols to achieve efficient data integration and process management. For personalized recommendation scenarios, the system develops precise task execution plans based on in-depth user profile analysis.
The LLM Planner is also equipped with professional task optimization mechanisms. Through systematic summarization of execution data, it continuously improves planning strategies. For example, through in-depth analysis of system call efficiency and task completion quality, it constantly optimizes the execution framework. This continuous improvement mechanism enables LLM Planner to demonstrate significant advantages in intelligent recommendations, automated management, and other areas, providing reliable technical support for complex business scenarios.

\subsection{Reflection and Refinement}
\subsubsection{Reflection Framework}
The reflection framework is an advanced learning mechanism that significantly improves model task execution through the self-reflection capabilities of language agents.
Language Reinforcement Learning: This framework transforms system feedback into structured language analysis and stores it in dedicated cache as reference for subsequent tasks. This approach enables the model to systematically analyze past experiences and continuously optimize decision paths.

Implementation Mechanism: The framework employs precise task attribution analysis to generate specific improvement plans. Through systematic evaluation methods and automated testing, the framework can accurately identify optimization opportunities and provide professional improvement suggestions.
Iterative Optimization: The framework adopts a learning pattern similar to human cognition, optimizing execution strategies through systematic analysis of historical data. In practical applications, the model can continuously adjust solutions based on test results to achieve performance improvements.
By integrating this professional reflection mechanism into core functionality, the framework demonstrates significant advantages across multiple application domains.

\subsubsection{Self-Optimization Mechanism}
Self-optimization framework employs advanced iterative optimization mechanisms to continuously improve model output quality. Specific implementations include:

\begin{itemize}
    \item \textbf{Two-Phase Process}. The self-optimization framework is divided into two core phases:
    \begin{enumerate}
        \item \textbf{Evaluation}. The system first conducts professional evaluation of initial output, ensuring output meets quality standards through multi-dimensional analysis.
        \item \textbf{Optimization}. Systematic improvements are made based on evaluation results until preset performance indicators are achieved.
    \end{enumerate}
    \item \textbf{Continuous Optimization}. The framework maintains complete optimization records in each iteration, building systematic knowledge accumulation to effectively avoid recurring issues.
    \item {Application Domains}. This framework demonstrates excellent performance in multiple professional fields:
    \begin{itemize}
        \item In code optimization, the system can identify performance bottlenecks and provide professional optimization solutions.
        \item In dialogue systems, the framework ensures output content accurately meets user requirements.
    \end{itemize}
\end{itemize}

This innovative approach achieves significant performance improvements through autonomous optimization mechanisms without requiring additional training resources.

\subsection{Memory-Augumented Planning}
\subsubsection{Memory Stream} 
Memory Stream is the core module for storing comprehensive experience records of agents, recording events described in natural language form, along with creation timestamps and last access timestamps. It records all events perceived by the agent through "Observation" and stores generated "Reflection" and "Planning" results in the same data structure. Memory Stream serves as long-term storage for agent behaviors, capable of dynamic updates and providing support for other modules.

\subsubsection{Reflection} 
The reflection module refines low-level information into high-level abstract thoughts by summarizing agent observations and memories. For example, an agent can generate high-level reflections like "passionate about music creation" through multiple observations. These reflections are organized in a tree structure, with abstract thoughts at the top level and basic observations at the bottom, influencing the agent's long-term behavioral logic and future decisions.

\subsubsection{Planning}
The planning module is responsible for generating future behavior plans for agents, including location, start time, and duration. Agents refine high-level overviews into hourly and minute-level sub-plans recursively, making behaviors more detailed and logical. The planning module can also dynamically adjust plans based on environmental changes, ensuring flexibility and consistency in agent behavior.

\subsubsection{ Agent Interaction}
The agent interaction module supports natural language dialogue between agents and real-time responses to the environment. Agents can generate dialogues based on memories and reflections, deepening their understanding of other agents through interaction. Additionally, agents can perceive changes in environmental states, such as stove burning, and take immediate action in response to these dynamic changes.

\subsubsection{Sandbox Environment}
The sandbox environment is a virtual world for agent activities, containing structured elements such as scenes, sub-scenes, and objects. Agents explore the sandbox environment, update their environment trees, and execute tasks. The sandbox environment provides concrete scenario support for agent behaviors, where task execution directly affects environmental states, demonstrating behavioral coherence and impact.

\subsubsection{Retrieval} 
The memory retrieval mechanism extracts the most relevant data for current tasks from the memory stream using strategies based on recency, importance, and relevance. Through semantic similarity calculations, agents can dynamically extract memories highly relevant to the current context, providing crucial support for planning, reaction, and decision-making. This mechanism ensures agents can quickly adapt to complex environments and make logical behavioral responses.

\subsection{Co-Training}
A simple combination of multiple models often fails to achieve the expected results. This primarily stems from each independent model's lack of deep understanding of other models' specialized domains, along with significant challenges in coordinating task objectives, execution standards, and contextual information. This situation constrains the collaborative efficiency between models, preventing them from fully leveraging their respective advantages.
To effectively address this challenge, we propose establishing a unified learning ecosystem. Through the integration of data resources and application scenarios, we implement a Co-Training collaborative training strategy, enabling models to undergo systematic learning and iterative optimization in a shared environment. This approach encourages different models to form complementary mechanisms when processing shared data, continuously improving their general performance and overall effectiveness through deep interaction. This systematic collaboration mechanism not only breaks through the limitations of single models but also generates significant synergistic benefits, achieving system performance improvements that exceed simple addition.

\subsection{Evaluation}
\subsubsection{Functional Achievement Assessment}
Systematic evaluation of whether models or agents achieve preset goals. Specifically, we assess whether their performance in completing web tasks meets expected requirements.
The evaluation process focuses on the precision of task completion status, such as verifying order processing status or the accuracy of inventory information updates.

\subsubsection{Multi-level Task Assessment}
Focuses on evaluating the execution of compound tasks involving multiple steps, especially in cross-platform operation scenarios. Using MIND2WEB and WebArena as examples, these platforms contain continuous tasks requiring precise execution of multiple stages.
WebArena emphasizes the success rate of state transitions throughout tasks and conducts in-depth analysis of diverse task completion paths.

\subsubsection{Adaptability Assessment} 
In-depth evaluation of model performance when facing new environments. For example, using MIND2WEB to assess system adaptability in handling new webpage architectures and task requirements.
Comprehensively evaluates the model's practical application capabilities through multi-dimensional scenario testing (including tourism, commerce, service, and other domains).

\subsubsection{ Logic Transparency Assessment}
In-depth analysis of the rationality of model decision processes. Using HOTPOTQA and FEVER as examples, these platforms provide detailed factual support to facilitate verification of the model's logical foundations. Systematically evaluates reasoning process accuracy by comparing model reasoning bases with standard answers.

\subsubsection{Comprehensive Reasoning Assessment}
Evaluates the system's ability to integrate multi-source information to complete complex tasks. HOTPOTQA particularly emphasizes the model's need for deep reasoning analysis from multiple information sources.

\subsubsection{Environmental Response Assessment}
In dynamic platforms like WebArena, focuses on evaluating system adaptability to real-time changes, including navigation efficiency, interaction quality, and exception handling capability.
Ensures that tasks completed by models or agents functionally achieve expected goals. For example, after completing a webpage task, checking whether results align with high-level intentions in the task description.
Specific operations include checking whether the final state of task execution matches expectations, such as verifying successful order placement or warehouse content updates.

\section{Joint Mining Mechanism}
Behavioral economics research shows that people's decisions are often influenced by subconscious biases, especially when assessing risks and rewards~\citep{Tversky-1974,Tversky-1991,Tversky-1992,Redelmeier-1996,chen2024decoy,chen2023reference,Kahneman-1991,Kahneman-2003,Kahneman-2011}. In our framework, by providing long-term incentives for early investments, we can effectively motivate participants to continually invest resources and innovate technologically. This incentive mechanism not only enhances the efficiency of model development but also ensures that the interests of both parties in the collaboration are maximized.

\begin{figure*}[htbp]
\includegraphics[width=.95\linewidth]{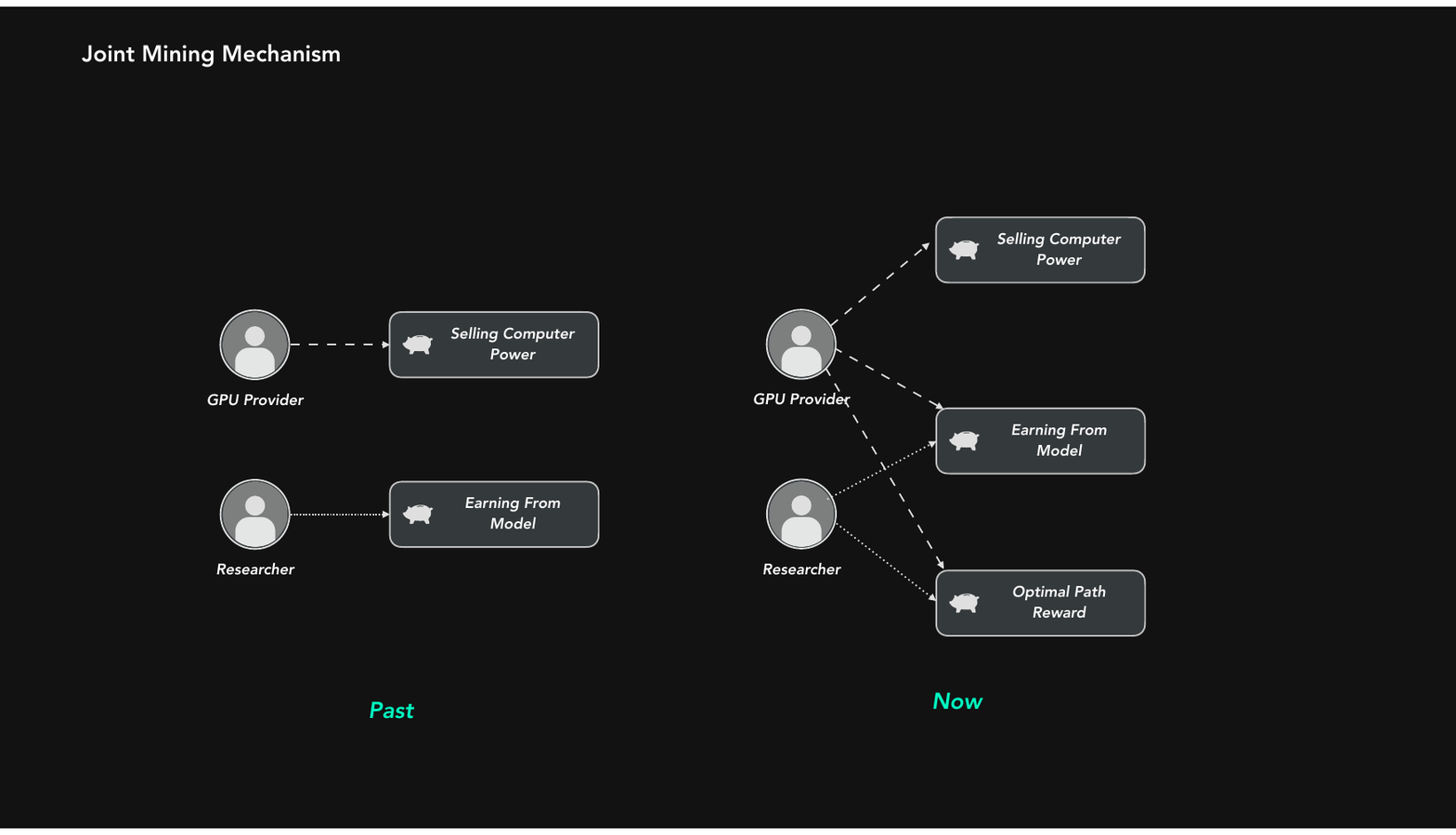}
  \caption{The illustration of the joint mining mechanism.}
  \Description{}
  \label{fig:dimensions}
\end{figure*}
\subsection{Computing Power Providers - Computing Power Investors}
In traditional computing power supply mechanisms, computing power providers mainly refer to GPU card providers. This role was previously more like a pure hardware supplier, but under our framework, computing power providers now have their own "operating system." This is similar to how IBM hardware once ran Microsoft's operating system. We hope that in the AI development cycle, hardware and computing power providers can truly inject soul into their contributions.
We have opened up this new model of computing power investment to help computing power providers transform short-term returns into long-term benefits. In traditional cooperation models, regardless of how innovative the model becomes or how many people use it, computing power providers rarely share in the benefits brought by the model. The joint mining mechanism effectively balances this inequality in the benefit cycle.
This model is actually quite similar to some marriage relationships: women need to invest significant time and energy early on, sacrificing career opportunities to take on the responsibilities of childbirth and raising the next generation; while men often only manage to give back to the family through promotions and salary increases many years later. Marriage is, to some extent, a mapping of the joint mining model—we need to ensure that the party who contributes more in the early stages can fairly share in the long-term benefits.

\subsection{Model Developers}
Model developers can be mainly divided into three categories: AI researchers from large enterprises, master's and doctoral students from university laboratories, and individual researchers. For these developers, not only do we need to lower technical barriers, but we also need to address the economic pressures during the training process. Especially for university laboratory students and individual researchers, computing resources are often very scarce, and the enormous economic burden frequently hinders their initial research.
The joint mining mechanism essentially provides model developers with more opportunities for trial and error. Researchers can build their designed models at lower costs without waiting for lengthy GPU queues. This is exactly the change we hope to bring in order to benefit the whole research community, as one interviewee argued that they deeply experienced the scarcity of GPU resources at their university, even though the university was the top-notch and was very supportive of research, resources were still limited. In return, we expect model designers to share some long-term benefits with computing power investors to thank them for their initial support and trust.

\subsection{Optimal Path Rewards}
In our design, the optimal path refers to selecting the path that best meets target requirements, has the highest efficiency, or lowest cost from multiple possible solutions through specific evaluation criteria and optimization algorithms. In large- scale complex systems, especially multi-level Large Language Model (LLM) combinations, the selection of optimal paths is particularly important as it directly affects system performance, resource utilization, and response speed. In building and optimizing multi-level LLM combination systems, we use components like LLM Planner and Optimal Path Evaluator to automatically combine multiple LLMs into a hierarchical tree structure, and automatically explore optimal combination paths through training and fine-tuning.
The Optimal Path Evaluator is responsible for evaluating and optimizing multiple combination paths generated by the LLM Planner. Through the collaborative work of these two components, the LLM system forms a hierarchical tree structure that both clearly defines the division of responsibilities among LLM levels and provides good scalability and maintainability for the system.
The optimal path is like a world record in the AI world, with each breakthrough representing a major innovation.

\subsection{Model Copyright Benefits}
In our model domain, copyright benefits mainly involve two roles: computing power investors and model designers. Through joint mining mechanisms and smart contracts, we have established a fair, transparent, and efficient copyright benefit distribution system to ensure reasonable returns for all parties in long-term cooperation.
Based on blockchain technology, all copyright benefit distribution records are permanently stored on-chain, allowing participants to view and verify distribution status at any time. This transparent mechanism effectively prevents unfair phenomena in benefit distribution and enhances mutual trust among parties.
Our copyright benefit system has multiple advantages: first, it protects the basic economic interests of model designers and computing power investors; second, through long-term cooperation mechanisms, benefits continue to grow as models are optimized and application scope expands, incentivizing stable cooperation among all parties; third, model designers can earn more copyright benefits through continuous innovation to enhance model value; finally, computing power investors receive continuous returns by providing stable and efficient computing power support to ensure high-performance model operation.

\bibliographystyle{ACM-Reference-Format}
\bibliography{sample-base}


\end{document}